 \documentclass[letterpaper, 10 pt, conference]{ieeeconf}       

\IEEEoverridecommandlockouts                  
                                                          
\usepackage{graphics} 
\DeclareGraphicsExtensions{.eps, .pdf, .jpeg, .jpg, .png}
\usepackage{graphicx}
\usepackage{ctable} 
\usepackage{bbm}
\usepackage{dsfont}
\usepackage{color}
\usepackage{multirow}
\usepackage{tabularx}
\usepackage{adjustbox}
\usepackage{amsmath}
\usepackage{url}
\usepackage[ruled,linesnumbered]{algorithm2e}
\usepackage{soul}
\usepackage{cite}
\usepackage{graphicx}
\usepackage{color,soul}

\title{\LARGE \bf 
Future Aspects in Human Action Recognition: \\Exploring Emerging Techniques and Ethical Influences
}

\author{Antonios Gasteratos$^{1}$, Stavros N. Moutsis$^{1,*}$, Konstantinos A. Tsintotas$^{1}$, and Yiannis Aloimonos$^{2}$ 
~
\thanks{$^{1}$Authors are with the Department of Production and Management Engineering, Democritus University of Thrace, 12 Vas. Sophias, GR-671 32, Xanthi, Greece {\tt\small $\lbrace$agaster, smoutsis, ktsintot$\rbrace$@pme.duth.gr}}
\thanks{$^{2}$Author is with the Department of Computer Science, University of Maryland, College Park, MD 20742, USA {\tt\small $\lbrace$yiannis$\rbrace$@umiacs.umd.edu}}
\thanks{$^{*}$Corresponding Author}
}
\graphicspath{{figures/}} 

\begin{document}

\maketitle
\thispagestyle{empty}
\pagestyle{empty}



\section{The Perspective of Human Action Recognition} 
\label{sec:introduction}

Visual-based human action recognition can be found in various application fields, \textit{e.g.,} the surveillance systems for safety and security~\cite{8591338}, sports analytics~\cite{9999033}, as well as medical assistive technologies~\cite{LAFUENTEARROYO2022116715}, or human-robot interaction frameworks~\cite{9956300}, and concerns the identification and classification of individuals' activities within a video.
Since actions typically occur over a sequence of consecutive images, it is particularly challenging due to the inclusion of temporal analysis~\cite{wang2016temporal}, which introduces an extra layer of complexity, in contrast to image classification and object detection tasks focusing solely on spatial pattern recognition.
However, even though multiple approaches try to handle temporal analysis~\cite{wang2016temporal, 6165309, ZHANG2020304, Arnab_2021_ICCV}, there are still difficulties because of their computational cost and lack of adaptability~\cite{Arnab_2021_ICCV}.
Therefore, different types of vision data, containing transition information between consecutive images, provided by next-generation hardware sensors will guide the robotics community in tackling the problem of human action recognition.
Using such data, researchers will be able to focus on processing the temporal data instead of attempting to capture the time domain in a video, increasing, this way, a system's performance.

On the other hand, while there is a plethora of still-image datasets~\cite{5206848}, that researchers can adopt to train new artificial intelligence models, videos representing human activities are of limited capabilities, \textit{e.g.,} small and unbalanced datasets or selected without control from multiple sources~\cite{singh2019video}.
The most important reason concerns their difficulty during selection and labeling, while due to their data capacity, more storage requirements are needed.
To this end, generating new and realistic synthetic videos is possible since labeling is performed throughout the data creation process.
On the contrary, emerging strategies that rely on reinforcement learning (RL)~\cite{OLIFF2020326} can permit the avoidance of considerable dataset dependence. 

At the same time, human factors' involvement raises ethical issues for the research community, as doubts and concerns about new technologies already exist.
These subjects are independent of the performance and data enhancement and should be carefully considered.
Hence, future pipelines led by ethical boundaries should provide moral awareness to improve society's trust~\cite{liang2021artificial, etemad2022ethical}.
As a reminder of this abstract paper, Section~\ref{sec:temp} answers the challenges of temporal analysis. Next, Section~\ref{sec:tools} gives directions towards tools and techniques capable of overcoming or enhancing datasets with limited properties, while the last section briefly discusses the ethical influences.

\section{Temporal Analysis}
\label{sec:temp}


Most common pipelines addressing human action recognition use pre-process strategies, such as optical flow and short-time motion energy~\cite{abdelbaky2020human}, which often are fused with the output of spatial analysis~\cite{NIPS2014_00ec53c4}.
However, another direction of approaches is based on the spatiotemporal analysis, \textit{e.g.}, 3 Dimension Convolutional Neural Networks (3D-CNNs)~\cite{6165309}, Convolutional Long Short-Term Memory~\cite{ZHANG2020304}, and Video Vision Transformers (ViViT)~\cite{Arnab_2021_ICCV}, where temporal and spatial study is performed simultaneously enhancing the overall performance.
Yet, the common characteristic of both categories is their high computational complexity, making them unsuitable for robotic applications~\cite{robotics12060167}.

\begin{figure}[t!]
    \begin{center}
    \includegraphics[width=7.5cm]{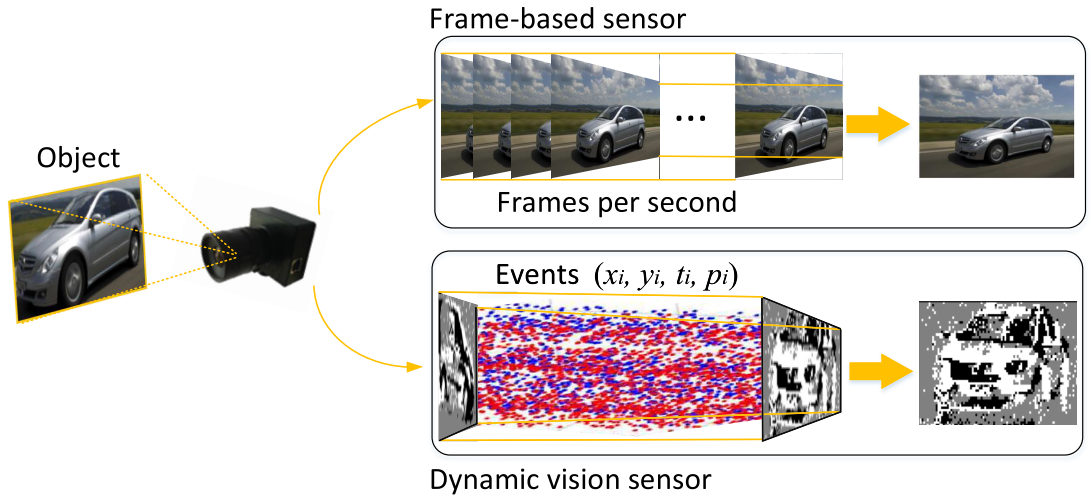}
    \centering
    \caption{Event cameras capture pixel-level changes in brightness with high temporal precision (with permission from~\cite{9023482}).
    }
    \label{fig:event}
    \end{center}
\end{figure}

Consequently, accurate systems require easily applicable, trainable, and accessible baseline approaches in temporal-based systems~\cite{Arnab_2021_ICCV}.
Some frameworks above possess these desirable characteristics, as they have released pre-trained networks that can be easily applied or fine-tuned. 
Nevertheless, they remain computationally expensive.
Therefore, the solution might lie in the type of data being processed rather than the techniques being used. 
Toward this goal, the integration of event cameras in human action recognition tasks~\cite{Arnab_2021_ICCV, 10198747}, whose output includes the dimension of time (see Fig.~\ref{fig:event}), could be an optimized solution for the temporal analysis.
This would allow for higher temporal performance and computational efficiency as the event data would be processed at the hardware level.
~

\section{Tools \& Techniques to Handle the Limited Datasets' Capabilities}
\label{sec:tools}

The available human action recognition video datasets are limited and smaller, often generated in specific environments and experienced actors.
To this end, they are unsuitable for general applications, as the creation of biased models is possible.
Therefore, it is necessary to develop tools and employ techniques that enable models to learn and, by extension, robots to behave and interact independently of specific datasets.
Hence, synthetic image sequences are generated for deep learning techniques~\cite{jimaging8110310} when data is insufficient.

While the creation of videos is a complex process, most recently, new tools have been proposed, such as Text-to-Video~\cite{singer2022makeavideotexttovideogenerationtextvideo} that have simplified the generation of synthetic video datasets.
This way, diverse data depicting specific actions can be generated and combined with existing human action recognition image sequences to create comprehensive datasets, leading to unbiased models.  

Moreover, techniques like RL, where models are trained with penalty-reward methods instead of datasets, may be a possible solution in daily actions.
However, for the extended use of RL in HRI in daily actions, simulated environments are necessary, like the ones created for the training of autonomous cars, \textit{e.g.}, the Car Learning to Act~\cite{MALIK2022742}.

\section{Ethical Influences}

\label{sec:ethical}

Systems that respect and follow ethical frameworks will be more adopted, as users will feel more secure and confident~\cite{liang2021artificial, etemad2022ethical}.
A key step to achieving this is to create pipelines that consider ethical issues and boundaries from the outset.
In addition, systems need to adapt to known concerns. 
Ethical matters are a socio-political phenomenon; by extension, they change according to place and time. 
Therefore, systems that will not have the ability to adapt to changes will most likely become inactive, as they will not be socially or legally acceptable.

Consequently, ethical considerations will largely guide the creation of artificial intelligence systems, while systems involving the human factor are of particular concern in moral discussions.
For instance, activities that would violate human personality to be recognized and recorded by robotic systems should be disregarded and prevented from being identified. 
However, ethical issues are not only socio-political but also personal. 
For one user, a process might violate their rights, while for another, it may be necessary for their safety and convenience.
Therefore, it would be helpful if the methods could be adapted according to the users' needs and requirements\cite{10759155}.

\bibliography{Bibliography}
\bibliographystyle{IEEEtran}

\end{document}